# Feature Level Clustering of Large Biometric Database

Hunny Mehrotra
NIT Rourkela
Rourkela 769008
hunny@nitrkl.ac.in

Dakshina R. Kisku
Javdavpur University
Kolkata 700032
drkisku@ieee.org

V. Bhawani Radhika
NIT Rourkela
Rourkela 769008
vbradhika27@gmail.com

Banshidhar Majhi
NIT Rourkela
Rourkela 769008
bmjahi@nitrkl.ac.in

Phalguni Gupta
IIT Kanpur
Kanpur 208016
pg@iitk.ac.in

**Abstract**

*This paper proposes an efficient technique for partitioning large biometric database during identification. In this technique feature vector which comprises of global and local descriptors extracted from offline signature are used by fuzzy clustering technique to partition the database. As biometric features posses no natural order of sorting, thus it is difficult to index them alphabetically or numerically. Hence, some supervised criteria is required to partition the search space. At the time of identification the fuzziness criterion is introduced to find the nearest clusters for declaring the identity of query sample. The system is tested using bin-miss rate and performs better in comparison to traditional k-means approach.*

## 1 Introduction

Existence of a large number of biometric records in the database requires rapid and efficient searching method. With the increase in the size of the biometric database, reliability and scalability issues become the bottleneck for low response time, high search and retrieval efficiency in addition to accuracy. Traditionally identification systems claims identity of an individual by searching templates of all users enrolled in the database [1]. These comparisons increase the data retrieval time along with the error rates. Thus a size reduction technique must be applied to reduce the search space and thus improve the efficiency. Conventionally databases are indexed numerically or alphabetically to increase the efficiency of retrieval. However, biometric databases do not posses a natural order of arrangement which negates the idea to index them alphabetically/numerically. Reduction of search space in biometric databases thus remains a challenging problem.

To reduce search space certain classification, clustering and indexing approaches have been proposed. In supervised classification or discriminant analysis, a collection of labeled (pre-classified) patterns are provided; the problem is to label a newly encountered, yet unlabeled, pattern. Typically, the given labeled (training) patterns are used to learn the descriptions of classes which in turn are used to label a new pattern. There exist several classification techniques like classification of face images based on age [2] where input images can be classified into one of three age-groups: babies, adults, and senior adults. Gender classification from frontal facial images using genetic feature subset selection is considered in [3]. Most of the existing fingerprint classification approaches make use of the orientation image [4]. An algorithm for the automatic coarse classification of iris images using box-counting method to estimate the fractal dimensions of iris is given in [5]. The main drawback of classification is that it is the supervised method where number of classes has to be known in advance. Further the data within each class is not uniformly distributed so the time required to search some classes is comparatively large.

The limitations of classification can be addressed with unsupervised approach known as ***Clustering***. It involves the task of dividing data points into homogeneous classes or clusters so that items in the same class are as similar as possible and items in different classes are as dissimilar as possible [6]. Intuitively it can be visualized as a form of data compression, where a large number of samples are converted into a small number of representative prototypes. Clustering can be broadly classified into Hard and Fuzzy approaches [7]. Non-fuzzy or hard clustering divides data into crisp clusters, where each data point belongs to exactly one cluster. Fuzzy clustering segments the data such that each sample data point can belong to more than one cluster and each data point has some degree of association with every cluster. The sum of the membership grades of a particular data point belonging to more than one cluster is always one.

From the available biometric features it has been inferred that each feature set has an association with more than one cluster and may have dissimilarity with data of the same cluster. In other words they are said to show inter class similarities and intra class variations, thus making them difficult to assign them to a single cluster. For example, variations in the face image of an individual due to change in pose, expression, lighting and eye glasses. Hence fuzzy clustering techniques prove to be an efficient means for grouping biometric data.

The paper proposes an efficient technique to segment large biometric database using fuzzy clustering. The concept of fuzzy clustering is discussed in Section 2. As a matter of case study the proposed technique has been applied on signature database. The features that are taken into consideration are discussed in Section 3.1. Further an approach for identification is presented in Section 3.2. Experimental results of the proposed system have been analyzed in Section 4. Conclusions are given in the last section.

## 2 Fuzzy C Means Clustering

Clustering involves the process of arranging data points in such a way that items sharing similar characteristics are grouped together. The goal is to find the natural grouping of data points without prior knowledge of class labels (unsupervised). Fuzzy C Means (FCM) is a feature clustering technique wherein each feature point belongs to a cluster by some degree that is specified by a membership grade [8]. These kind of clustering algorithms are known as objective function based clustering. Given $M$ dimensional database of size $N$ where $N$ is the total number of feature vectors and $M$ is the dimension of each feature vector. FCM assigns every feature vector a membership grade for each cluster. The problem is to partition the database based on some fuzziness criteria using membership values. To find membership values, the partition matrix $U$ of size $N \times c$ is calculated that defines membership degrees of each feature vector. The values 0 and 1 in $U$ indicate no membership and full membership respectively. Grades between 0 and 1 indicate that the feature point has partial membership in a cluster. The following steps are involved in training the database using FCM technique

### 2.1 Initialization of the partition matrix

Initially a fuzzy partition matrix $U$ is generated that is of size $N \times c$, where $c$ is number of clusters and $N$ is total number of feature vectors. Subject to the constraint that

$$\sum_{j=1}^{c} U_{ij} = 1, \quad \forall i \varepsilon \{1,2,...N\} \quad (1)$$

### 2.2 Calculation of fuzzy centers

The fuzzy centers are calculated using the partition matrix generated in 2.1.

$$C_j = \frac{\sum_{i=1}^{N} U_{ij}^m \times x_i}{\sum_{i=1}^{N} U_{ij}^m} \quad (2)$$

where $m \geq 1$ is a fuzzification exponent. The larger the value of $m$ the fuzzier the solution will be. This indicates the number of iterations that is required for clustering. $x_i$ is $i^{th}$ feature vector. The value of $i$ ranges from 1 to $N$ (total number of templates in the database).

### 2.3 Updating membership and cluster centers

FCM is an iteration loop. The method of clustering is based on minimization of the objective function defined by

$$J_m = \sum_{i=1}^{N} \sum_{j=1}^{C} U_{ij}^m \|x_i - c_j\|^2 \quad (3)$$

$U_{ij}$ describes the degree of member of feature set ($x_i$) with cluster $c_j$. $\|*\|$ represents norm between $x_i$ and cluster center $c_j$ given by

$$\| x_i - c_j \|^2 = (x_i - c_j)^T A(x_i - c_j) \quad (4)$$

where $A$ is identity matrix for Euclidean distance used here. At every iteration the membership matrix is updated using

$$U_{ij} = \frac{1}{\sum_{k=1}^{c} \left( \frac{\|x_i - c_j\|}{\|x_i - c_k\|} \right)^{\frac{2}{m-1}}} \quad (5)$$

The revised membership matrix (generated in (5)) is used for updating the cluster centers using equation (2). The iteration will stop when $max_{ij}\{|U_{ij}^{(m+1)} - U_{ij}^{(m)}|\} < \varepsilon$, where $\varepsilon$ is a termination criteria. The value of $\varepsilon$ ranges between 0 and 1.

---

**Algorithm:** *fcmcluster (c: no of clusters, x: input data, N: total number of training data)*

**Step 1:** Fix $1 \leq m < \infty$, initial partition matrix $U^0$ (N×c), and the termination criterion $\varepsilon$.

**Step 2:** Calculate the fuzzy cluster centers c using equation (2).

**Step 3:** Update membership matrix as per equation (5).

**Step 4:** Calculate change in membership matrix $\Delta = \| U^{m+1} - U^m \| = max_{ij}|U_{ij}^{m+1} - U_{ij}^m|$. If $\Delta > \varepsilon$, then set m=m+1 and go to step 2. If $\Delta \leq \varepsilon$, then stop.

---

## 3 Signature Biometrics as a Case Study

As a case study the methodology discussed in Section 2 is applied to partition the large biometric database comprising of signature features. The steps involved in clustering the signature database are given as below:

### 3.1 Feature extraction and training

Signature is a behavioral characteristic [9] of a person and can be used to identify/verify a person's identity. The signature recognition algorithm consists of two major modules i.e., preprocessing and noise removal and feature extraction. Offline signature acquisition is carried out statically, unlike online signature acquisition, by capturing the signature image using a high resolution scanner. A scanned signature image may require morphological operations like normalization, noise removal by eliminating extra dots from the image, conversion to grayscale, thinning and extraction of high pressure region. The features of the signature images can be classified into two categories: global and local [10].

#### 3.1.1 Global features

Global features include the global characteristics of an image. Ismail and Gad [9] have described global features as characteristics which identify or describe the signature as a whole. Examples include: width/height (or length), baseline, area of black pixels etc. They are less responsive to small distortions and hence are less sensitive to noise as well, compared to local features which are confined to a limited portion of the signature.

#### 3.1.2 Local features

Local features in contrast to global features are susceptible to small distortions like dirt but are not influenced by other regions of the signature. Hence, though extraction of local features requires a huge number of computations, they are much more precise. However, the grid size has to be chosen very carefully. It can neither be too gross nor too detailed. Examples include local gra-dients, pixel distribution in local segments etc. Many of the global features such as global baseline, center of gravity, and distribution of black pixels have their local counterparts as well. The features obtained from an input signature image are listed as follows:

1. Width to height ratio

2, 3. Center of gravity (both X and Y coordinates) to height ratio

4. Normalized area of black pixels

5. Total number of components of the signature

6. Global Baseline to height ratio

7. Upper extension to height ratio

8. Lower extension to height ratio.

9, 10. Center of gravity (both X and Y coordinates) of the HPR image to height ratio

11. Area of black pixels in the HPR image to total area of black pixels in the image.

12. Number of cross points to area of black pixels in the thinned image

13. Number of edge points to area of black pixels in the thinned image

14. Slope of the thinned image

15. Trace to area of black pixels in the thinned image

16 to 27. Ratio of centre of gravity co-ordinates to height, ratio of pixel count of individual sections to total pixel count of the image and ratio of baseline position to height of the image in the 3 horizontal sections.

The feature set comprises of

$$F_i = [f_1 \ f_2 \ldots\ldots\ldots f_{27}] \quad (6)$$

where *i* ranges from 1 to *N* (total number of templates in the database). The features extracted are used for partitioning the database using FCM clustering technique given in Section 2. At the time of training each data item ($F_i$ with 27 values) is used to find the membership grade with every cluster center. Data is assigned to cluster with highest value of membership.

### 3.2 Identification strategy

The identification technique takes into consideration the membership matrix and finds the nearest cluster. Given a query data $q=[q_1 \ q_2 \ q_3 \ldots q_M]$ the approach updates the membership matrix using exponential modification. Further the Euclidean distance between the $j^{th}$ cluster centre *c* and query data *q* is obtained using

$$dist(j) = \sqrt{(q-c_j)^2} \quad (7)$$

After obtaining the distance with each cluster centre the objective function is calculated as given in equation (3) using initial membership matrix. The membership matrix is updated using calculated distance values (equation (7)) as given in equation (5). The updated membership matrix is checked for termination criteria against *ε*. If criteria is met the iteration stops. The fuzzy factor is brought into consideration by choosing clusters with two maximum values of membership grades. The retrieved clusters are chosen to be target clusters to find suitable matches for a particular query signature. The selected templates (*K*) corresponding to the target cluster ($K \subseteq N$) are retrieved from the database and compared to query template to find a match. The system diagram of proposed identification technique is shown in Figure 1. This technique is a preferred over hard clustering techniques as more than one cluster is taken into consideration to declare the identity of an individual. The algorithm for identification is given as follows

---

**Algorithm:** *identify (q: query data, c: cluster centres)*

**Step 1:** Calculate distance dist between q and c. Initialize the partition matrix $U^m$.

**Step 2:** Update the partition matrix $U^{m+1}$ by using dist and $U^m$.

**Step 3:** Calculate change in partition matrix $\Delta = \| U^{m+1} - U^m \| = \max_{ij} U_{ij}^{m+1} - U_{ij}^m$. If $\Delta > \varepsilon$, then set m = m + 1 and go to step 2. If $\Delta \leq \varepsilon$, then stop.

**Step 4:** Find two max$\{U^{m+1}\}$ and retrieve target clusters.

---

## 4 Experimental Results

The results are obtained on signature database collected by the authors. The database comprises of signatures from 1000 individuals. Each individual gives nine signatures on a custom defined template. The user is asked to sign within a box. Among the nine signatures available, first six signatures are used for enrollment and last three are used for searching and identification. To measure the performance of the system, bin-miss rate is obtained by varying the number of clusters as shown in Figure 2. Bin-miss rate gives the number data that has not fallen into proper cluster. From the graph it is evident that the bin-miss rate increases with increase in the number of clusters (*c*). This implies that by taking two neighboring clusters in case of FCM, poorly whole database is searched for *c* equal to 2. So an optimum value of *c* is required that gives good accuracy with large partitioning of sample space. The comparative study is presented in graph as well as Table 1. From the Table it is evident that when number of clusters is less K-Means performs better as compared to FCM. The reason underlying this is that the hard clustering approaches performs better when database is divided into less number of clusters. However as the number of cluster increases the probability of data lying in a proper cluster becomes very low. Thus use of fuzzy criteria helps in minimizing errors. Here membership grade with pre-computed cluster centers acts as fuzzy criteria.

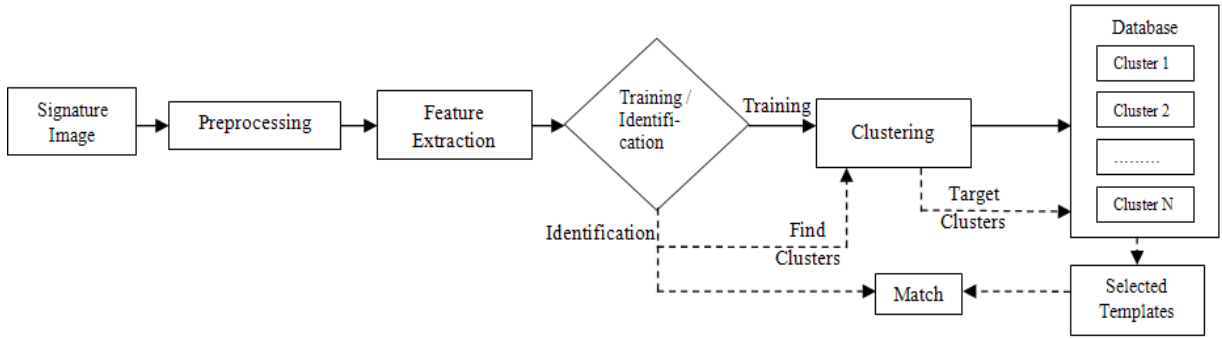

Figure 1 System diagram of clustering based identification technique

Table 1. Bin Miss Rate for different clusters using FCM and K-Means

| No of clusters | FCM | K-Means |
|---|---|---|
| 2 | 1 | 0 |
| 3 | 2 | 0 |
| 4 | 3 | 1 |
| 5 | 8 | 8 |
| 6 | 11 | 12 |
| 7 | 12 | 18 |
| 8 | 16 | 21 |
| 9 | 17 | 25 |

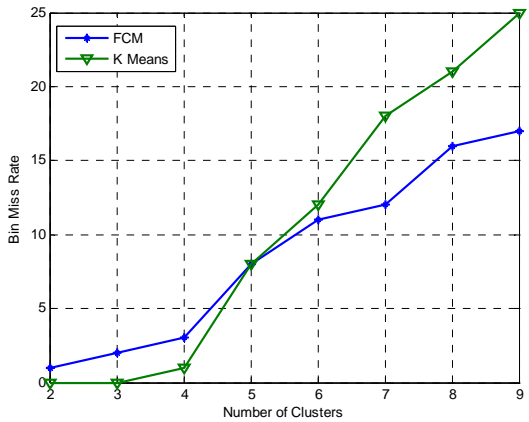

Figure 2 Graph showing bin miss rate by varying number of clusters for FCM and K-Means

## 5 Conclusion

This is an efficient approach to partition the large biometric database, to reduce data retrieval time during identification. The limitations of hard clustering techniques have been removed by introducing the fuzziness criteria. Here fuzziness factor is essential owing to the nature of biometric database. The system is performing comparatively superior as compared to traditional K-Means clustering technique. For less number of clusters the approach is not suitable. However as the size of database increases the number of clusters required for partitioning also increases. Thus it is a preferred partitioning technique for large scale biometric systems. There is still scope of research to find optimum number of clusters that can give maximum accuracy with reduced size of search space for the matcher.